\pdfoutput=1
\documentclass{article}
\usepackage{amsmath,graphicx,mlspconf}

\usepackage{epsfig}
\usepackage{calc}
\usepackage{amssymb}
\usepackage{amstext}
\usepackage{amsthm}
\usepackage{multicol}
\usepackage{pslatex}
\usepackage{multirow}
\usepackage{color}
\usepackage[ruled,linesnumbered]{algorithm2e}
\usepackage{algpseudocode}
\usepackage{tikz}
\usepackage{pgfplots}
\usetikzlibrary{spy}
\tikzset{annot/.style={draw=black,fill=white,text=black}}
\usepackage[caption=false]{subfig}

\usepackage[hyphens]{url}

\newcommand{\bbeta}{\mbox{\boldmath  $\beta$}}

\newcommand{\bSigma}{\mbox{\boldmath $\Sigma$}}

\newcommand{\bmu}{\mbox{\boldmath   $\mu$}}

\newcommand{\Ymat}{{\bf Y}}
\newcommand{\Zmat}{{\bf Z}}
\newcommand{\Bmat}{{\bf B}}
\newcommand{\Mmat}{{\bf M}}
\newcommand{\Nmat}{{\bf N}}
\newcommand{\Hmat}{{\bf H}}
\newcommand{\Rmat}{{\bf R}}
\newcommand{\Amat}{{\bf A}}
\newcommand{\Emat}{{\bf E}}

\newcommand{\Pmat}{{\bf P}}

\newcommand{\Vmat}{{\bf V}}

\newcommand{\Imat}{{\bf I}}
\newcommand{\Xmat}{{\bf X}}
\newcommand{\Cmat}{{\bf C}}
\newcommand{\Dmat}{{\bf D}}
\newcommand{\Wmat}{{\bf W}}
\newcommand{\bx}{{\bf x}}

\newcommand{\bv}{{\bf v}}

\newcommand{\bn}{{\bf n}}
\newcommand{\bP}{{\bf P}}

\newcommand{\by}{{\bf y}}
\newcommand{\bu}{{\bf u}}

\newcommand{\bF}{{\bf F}}
\newcommand{\bU}{{\bf U}}

\DeclareMathOperator*{\argmin}{argmin}

\newtheorem{theorem}{Theorem}
\newtheorem{lemma}{Lemma}
\title{Scene-adapted plug-and-play algorithm with convergence guarantees}
\name{Afonso M. Teodoro \quad Jos\'{e} M. Bioucas-Dias \quad M\'{a}rio A. T. Figueiredo\thanks{This work was partially supported by the {\it Funda\c{c}\~ao para a Ci\^encia e Tecnologia} (FCT), grants UID/EEA/5008/2013, BD/102715/2014, and PTDC/EEI-PRO/0426/2014.}}
\address{Instituto de Telecomunica\c{c}\~{o}es\\ Instituto Superior T\'{e}cnico, Universidade de Lisboa, Portugal }

\begin{document}
	\ninept
	\maketitle
	\begin{abstract}
		Recent frameworks, such as the so-called \textit{plug-and-play}, allow us to leverage the developments in image denoising to tackle other, and more involved, problems in image processing. As the name suggests, state-of-the-art denoisers are \textit{plugged} into an iterative algorithm that alternates between a denoising step and the inversion of the observation operator. While these tools offer flexibility, the convergence of the resulting algorithm may be difficult to analyse. In this paper, we plug a state-of-the-art denoiser, based on a Gaussian mixture model, in the iterations of an alternating direction method of multipliers and prove the algorithm is guaranteed to converge. Moreover, we build upon the concept of scene-adapted priors where we learn a model targeted to a specific scene being imaged, and apply the proposed method to address the hyperspectral sharpening problem.
	\end{abstract}
	
	\section{Introduction}
	\label{sec:intro}
	
	Image denoising is not only one of the core problems in image processing, but also a building block for many other tasks, and has been a very active research topic over the last decades. Most of the current state-of-the-art methods are patch-based, {\it i.e.}, they rely on a \textit{divide and conquer} principle, where instead of dealing with the noisy image as a whole, small (overlapping) patches are extracted, denoised independently, and put back in their locations \cite{dabov,Teodoro2015,ZoranWeiss}. Applying these patch-based methods in more general restoration/reconstruction problems is not trivial and has been a topic of research in the past years \cite{danielyan,Mignotte,Papyan2016}.
	
	Recently, Venkatakrishnan \textit{et al.} proposed a flexible framework, called \textit{plug-and-play} (PnP), where a state-of-the-art denoiser is seen as a black box and \textit{plugged} into the iterations of an \textit{alternating direction method of multipliers} (ADMM) \cite{Venkatakrishnan}. This approach allows using an arbitrary denoiser as a regularizer in an imaging inverse problem, such as deblurring or super-resolution, tackled via an ADMM algorithm, but departing from its standard use, where the denoiser is the proximity operator of some convex regularizer \cite{AADMM}. However, plugging an arbitrary denoiser (possibly without a closed-form expression) into ADMM begs obvious questions \cite{chan,Venkatakrishnan}: is the resulting algorithm guaranteed to converge? If so, does it converge to a (global or local) optimum of some objective function? Can we identify this function? Here, we give positive answers to these questions, when the plugged denoiser is a modified \textit{Gaussian mixture model} (GMM) based denoiser \cite{Teodoro2015,ZoranWeiss,yu}.
	
	
	
	
	As proposed in earlier work \cite{Teodoro2016}, the GMM can be adapted to specific classes of images; the rationale is that denoisers based on such class-adapted priors are able to better capture the characteristics of the class than a general-purpose denoiser. In this paper, we take this adaptation one step further, by considering scene-adapted priors. As the name suggests, the model is no longer learned from a set of images from the same class as the input image, but from one or more images over the same scene being restored/reconstructed. Hyperspectral sharpening \cite{review, simoes, qwei, Teodoro2017} and image deblurring with noisy/blurred image pairs \cite{yuan} are two examples of applications that may leverage from such priors.
	
	In summary, the contributions are threefold: (i) we prove that the ADMM algorithm with a plugged-in GMM-based denoiser is guaranteed to converge to the global minimum of a cost function, provided that we introduce a small modification to the denoiser; (ii) this proposed modification not only guarantees convergence of the algorithm, but it is also motivated by the scene-adapted perspective; (iii) we show that the proposed algorithm yields results that are often better than other state-of-the-art hyperspectral sharpening methods.
	
	The paper is organized as follows. Section~\ref{sec:admm} gives a brief description of ADMM, which is the basis of the proposed algorithm. Section~\ref{sec:pnp} explains the rationale behind the PnP framework, and our reasons for choosing the GMM-based denoiser. Section~\ref{sec:conv} contains what is arguably the main contribution of this paper: the convergence of the resulting algorithm. The application to HS sharpening is described in Section~\ref{sec:hs}, and experimental results are reported in Section~\ref{sec:results}. Finally, Section~\ref{sec:conclusion} concludes the paper.
	
	

	\section{Alternating Direction Method of Multipliers}\label{sec:admm}
	Although dating back to the 1970's \cite{gabay}, ADMM has seen a surge of interest in the last decade, as a flexible and efficient optimization tool, widely used in imaging problems, machine learning, and other areas \cite{eckstein, boyd}. One of the canonical problems for ADMM has the form
	\begin{equation}
	\min_\bx \;  f( \bx) + g(\Hmat  \bx), \label{eq:admmcan} \\
	\end{equation}
	where  $f$ and $g$ are closed, proper, convex functions, and matrix $\Hmat$ is of appropriate dimensions \cite{eckstein}. Each iteration of ADMM for \eqref{eq:admmcan} is as follows (with the superscript $(\cdot)^{(k)}$ denoting the iteration counter):
	\begin{align}
	\bx^{(k+1)} & =  \arg\min_\bx  f(\bx) + \frac{\rho}{2} \bigl\| \Hmat \bx -\bv^{(k)} - \bu^{(k)}\bigr\|_2^2   \label{eq:ineq1}\\
	\bv^{(k+1)} & =  \arg\min_\bv   g(\bv) + \frac{\rho}{2} \bigl\| \Hmat \bx^{(k+1)} - \bv -  \bu^{(k)}\bigr\|_2^2 , \label{eq:ineq2}\\
	\bu^{(k+1)} & =  \bu^{(k)} -  \Hmat \bx^{(k+1)} + \bv^{(k+1)} ,  \label{eq:ineq3}
	\end{align}
	where $\bu^{(k)}$ are the (scaled) Lagrange multipliers at iteration $k$, and $\rho$ is the penalty parameter.
	
	ADMM  can be directly applied to problems involving the sum of $J$ closed, proper, convex functions, composed with linear operators,
	\begin{equation}
	\min_{\bx } \sum_{j=1}^J g_j (\Hmat_{j}\bx) \label{eq:sum1J}
	\end{equation}
	by casting it into the form \eqref{eq:admmcan} as follows: 
	\begin{equation*}
	f(\bx) = 0, \hspace{0.25cm} \Hmat = \begin{bmatrix} \Hmat_{1} \\ \vdots \\ \Hmat_{J} \end{bmatrix}, \hspace{0.25cm} \bv = \begin{bmatrix} \bv_1 \\ \vdots \\ \bv_{J} \end{bmatrix}, \hspace{0.25cm} g(\bv) =  \sum_{j=1}^J g_j (\bv_j).
	\end{equation*}
	The resulting instance of ADMM (called SALSA--\textit{split augmented Lagrangian shrinkage algorithm} \cite{afonso2011augmented}) is
	\begin{eqnarray}
	\bx^{(k+1)} & =  &\arg\min_\bx  \sum_{j=1}^J \|  \Hmat_j \bx - \bv_j^{(k)}  -\bu_j^{(k)}  \|_2^2  \label{eq:ineq1b}\\
	\bv_1^{(k+1)} & =  &\arg\min_\bv   g_1(\bv) + \frac{\rho}{2} \bigl\| \Hmat_1 \bx^{(k+1)} - \bv -  \bu^{(k)}\bigr\|_2^2 , \label{eq:ineq2b}\\
	\vdots  & &  \vdots \nonumber \\
	\bv_J^{(k+1)} & = & \arg\min_\bv   g_J(\bv) + \frac{\rho}{2} \bigl\| \Hmat_J \bx^{(k+1)} - \bv -  \bu^{(k)}\bigr\|_2^2 , \label{eq:ineq2c}\\
	\bu^{(k+1)} & =  &\bu^{k} - \Hmat \bx^{(k+1)} + \bv^{(k+1)}, \label{eq:ineq3b}
	\end{eqnarray}
	where (as for $\bv^{(k)}$), $\bu^{(k)} = \bigl[ \bigl(\bu_1^{(k)}\bigr)^T, \, \dots, \, \bigl(\bu_J^{(k)}\bigr)^T\bigr]^T$.
	
	\section{Plug-and-Play Priors}\label{sec:pnp}
	The PnP framework \cite{Venkatakrishnan} emerges from noticing that sub-problems \eqref{eq:ineq2b}-\eqref{eq:ineq2c} can be interpreted as the MAP solutions to pure denoising problems. In such situations, the resulting sub-problem has an identity observation operator, function $g_i$ acting as the regularizer (or negative log-prior, in Bayesian terms), $(\bx^{(k+1)}\! - \bu^{(k)})$ as the noisy data, and noise variance equal to $1/\rho$. The goal is to capitalize on recent developments in image denoising, by using a state-of-the-art denoiser plugged into the iterations of ADMM, instead of trying to come up with a convex regularizer for which it is possible to compute the corresponding proximity operator.

	\subsection{GMM-Based Denoising}\label{ssec:gmmden}
	Amongst the several state-of-the-art denoising methods, we adopt a GMM patch-based denoiser. The reasons for this choice are threefold: (i) it has been shown that GMM are good priors for clean image patches \cite{ZoranWeiss, yu}; (ii) a GMM can be learned either from an external dataset of clean images \cite{ZoranWeiss, yu}, or directly from noisy patches \cite{Teodoro2015}; (iii) the ease of learning a GMM  from an external dataset  opens the door to  class-specific priors,  which, naturally, lead to superior results since they capture better the characteristics of the image class in hand, than a general-purpose image model \cite{Teodoro2016,luo}.
	
	
	
	Leaving aside for now the question of how to learn the parameters of the model,  we consider a GMM prior\footnote{$\mathcal{N}(\bx ; \bmu,{\bf C})$ denotes a Gaussian probability density function of mean $\bmu$ and covariance $\bf C$, computed at $\bx$.}
	\begin{equation}
	p({\bf x}_i ) = \sum_{j=1}^K \alpha_j \mathcal{N}({\bf x}_i ; \bmu_j, {\bf C}_j),
	\end{equation}
	for any patch ${\bf x}_i$ of the image to be denoised. A current practice in patch-based image denoising is to treat the average of each patch separately: the average of the noisy patch is subtracted from it, the resulting zero-mean patch is denoised, and the average is added back to the estimated patch. The implication of this procedure is that, without loss of generality, we can assume that the  components of the GMM have  zero mean ($\bmu_j = 0$, for $j=1,...,K$). 
	
	Given this GMM prior and a noisy version $\by_i$ of each patch, where $\by_i = \bx_i + \bn_i$, with $\bn_i \sim \mathcal{N}(0,\sigma^2)$, the \textit{minimum mean squared error} (MMSE) estimate of $\bx_i$ is given by
	\begin{equation}
	\hat{\bf x}_i = \sum_{j=1}^K \beta_j({\bf y}_i ) \; {\bf v}_j({\bf y}_i),\label{eq:MMSE}
	\end{equation}
	where
	\begin{equation}
	{\bf v}_j ({\bf y}_i) =  \Cmat_j \Bigl( {\bf C}_j +  \sigma^2 {\bf I} \Bigr)^{-1}  {\bf y}_i, \label{eq:v_m}
	\end{equation}
	and 
	\begin{equation}
	\beta_j({\bf y}_i) = \frac{\alpha_j \; \mathcal{N}({\bf y}_i ; 0, {\bf C}_j + \sigma^2 \, {\bf I})
	}{\sum_{k=1}^K \alpha_j \; \mathcal{N}({\bf y}_i ; 0, {\bf C}_k + \sigma^2 \, {\bf I})}. \label{eq:beta}
	\end{equation}
	Notice that  $\beta_j({\bf y}_i)$ is the posterior probability that the \textit{i}-th patch was generated by the \textit{j}-th component of the GMM, and ${\bf v}_j({\bf y}_i)$ is the MMSE estimate of the \textit{i}-th patch if it was known that it had been generated by the $j$-th component.

	As is standard in patch-based denoising, after computing the MMSE estimates of the patches, they are returned to their location and combined by straight averaging. This corresponds to solving the following optimization problem
	\begin{align}
	\widehat{\bx }  \in  \underset{\bx \in \mathbb{R}^n }{\argmin}  \quad  \sum_{i = 1}^{N} \Vert \hat{\bf x}_i   - \bP_i \bx \Vert_2^2, \label{eq:wholeopt}
	\end{align}
	where $\bP_i \in \{0,1\}^{n_p \times n}$ is a binary matrix that extracts the $i$-th patch from the image (thus $\bP_i^T$ puts the patch back into its place), $N$ is the number of patches, $n_p$ is the number of pixels in each patch, and $n$ the total number of image pixels. The solution to \eqref{eq:wholeopt} is
	\begin{equation}
	\hat{\bx} = \Bigl( \sum_{i = 1}^{N} \bP_i^T \bP_i \Bigr)^{-1}\Bigl( \sum_{i = 1}^{N} \bP_i^T \hat{\bf x}_i \Bigr) = \frac{1}{n_p}\,  \sum_{i = 1}^{N} \bP_i^T \hat{\bf x}_i, \label{eq:wholex}
	\end{equation}
	assuming the patches are extracted with unit stride and periodic boundary conditions, thus every pixel belongs to $n_p$ patches and $\sum_{i = 1}^{N} \bP_i^T \bP_i = n_p \Imat$.

	\subsection{From Class-Adapted to Scene-Adapted Priors}\label{ssec:class}
	Class-adapted priors have been used successfully in image deblurring, if the image to be deblurred is known to belong to a certain class, {\it e.g.}, image of a face or fingerprint, and a prior adapted to that class is used \cite{Teodoro2016}.
	Here, we propose taking this adaptation to an extreme, whenever we have two types of data describing the same scene, as is the case with data fusion problems. Instead of leveraging a dataset of images from the same class as the input image, we should leverage one type of data to learn a model that is targeted to the specific scene being imaged, and use it as a prior in the reconstruction/restoration process. The underlying assumption is that the two types of data share the same spatial statistical properties; this is a reasonable assumption, since the scene being imaged is exactly the same. In Section~\ref{sec:results}, we implement this idea to tackle a data fusion problem known as hyperspectral (HS) sharpening (see \cite{review} for a detailed description of the problem). We resort to a GMM prior, as described in Subsection~\ref{ssec:gmmden}, learned from (patches of the bands of) the observed multispectral (MS) or panchromatic (PAN) image using the classical \textit{expectation-maximization} (EM) algorithm, and then leverage this prior to sharpen the low resolution HS bands.
	
	In light of the scene-adaptation scheme just described, we further claim that, it not only makes sense to use a GMM prior learned from the observed MS or PAN image, but it also makes sense to keep the weights $\beta_m$ of each patch in the target image, used to compute \eqref{eq:MMSE}, equal to its value for the corresponding observed patch, obtained during training.  In other words, we use the weights as if we were simply denoising the high resolution MS image, with a GMM trained from its noisy patches. In fact, another instance of this idea has been previously used, based on sparse representations on learned dictionaries \cite{qwei}. We stress that this approach is only reasonable in the case of image denoising or image reconstruction using scene-adapted priors, where the training and target images are of the same scene, and thus have similar spatial structures. As will be shown in the next section, this modification will play an important role in proving convergence of the proposed PnP-ADMM algorithm.

	\section{Convergence of the {PnP-ADMM} Algorithm}\label{sec:conv}
	
	We begin by presenting a simplified version of a classical theorem on the convergence of a generalized version of ADMM, proved in the seminal paper by Eckstein and Bertsekas \cite{eckstein}. The full version of the theorem allows for inexact solution of the optimization problems in the iterations of ADMM, while this version, based on which our proof of convergence of PnP-ADMM will be supported, assumes exact solutions. 
	
	\begin{theorem}[Eckstein and Bertsekas \cite{eckstein}]\label{th:admm}
		Consider a problem of the form \eqref{eq:admmcan}, where $\Hmat$ has full column rank, and $f:\mathbb{R}^n\rightarrow \bar{\mathbb{R}}$ and $g:\mathbb{R}^m \rightarrow \bar{\mathbb{R}}$ are closed, proper, and convex functions, and let $\bv_0, \bu_0 \in \mathbb{R}^m$, and $\rho > 0$ be given. If the sequences $(\bx^{(k)}, k = 0, 1,\dots )$, $(\bv^{(k)}, k = 0, 1,\dots )$, and $(\bu^{(k)}, k = 0, 1,\dots )$ are generated according to  \eqref{eq:ineq1}, \eqref{eq:ineq2}, \eqref{eq:ineq3}, then  $(\bx^{(k)}, k = 0, 1,\dots )$ converges to a solution of \eqref{eq:admmcan}, $\bx^{(k)} \rightarrow \bx^{*}$, if one exists. Furthermore, if a solution does not exist, then at least one of the sequences $(\bv^{(k)}, k = 0, 1,\dots )$ or $(\bu^{(k)}, k = 0, 1,\dots )$ diverges.
	\end{theorem}
	
	Before proceeding, some properties of the denoising function are established.
	
	\subsection{Analysis of GMM-Based Denoiser}
	
	Consider the role of the denoiser: for a noisy input argument $\by \in \mathbb{R}^{n}$, it produces an estimate $\hat{\bx} \in \mathbb{R}^{n}$, given the parameters of the GMM, $\theta = \{\Cmat_1, \dots, \Cmat_K, \beta_1^1, \dots, \beta_K^n\}$, and a noise level, $\sigma$. As the GMM-denoiser is patch-based, the algorithm starts by extracting (overlapping, with unit stride) patches, followed by computing their estimates using \eqref{eq:MMSE}. Recall that, with the scene-adaptation scheme, parameters $\beta$ no longer depend on the noisy patches, and thus we may write compactly, for each patch,
	\begin{align}
	\label{eq:linMMSE}
	\hat{\bf x}_i &= \sum_{m=1}^K \beta_m^{i} \; {\bf C}_m \Bigl(  {\bf C}_m + {\sigma^2\, \bf I} \Bigr)^{-1} \by_i
	= \bF_i \, \by_i = \bF_i \, \bP_i\, \by,
	\end{align}
	where $\bP_i \in \{0,1\}^{n_p \times n}$ is the binary matrix defined in Subsection~\ref{ssec:gmmden}, thus $\by_i = \bP_i\, \by$. Moreover, we can relate the input of the denoiser to its image-wise output using \eqref{eq:wholex}, under the assumption of periodic boundary conditions. Once again, using compact notation
	\begin{equation}
	\hat{\bx} = \underbrace{\frac{1}{n_p} \sum_{i=1}^N \Pmat_i^T   \bF_i \Pmat_i}_{\Wmat} \; {\bf y} = \Wmat  \;  {\bf y}. \label{eq:defW}
	\end{equation}
	which shows that the GMM-based MMSE denoiser, with fixed weights, is a linear function of the noisy image. Matrix $\Wmat$, of course, depends on the parameters of the method ($\sigma$, $n_p$) as well as on the parameters of the GMM, but it is fixed throughout the iterations of the ADMM loop. Recall that the weights, $\beta$, and covariance matrices, $\Cmat$, are obtained during training, and $\sigma$ is assumed to be known, thus is fixed. 
	The next lemma states the key properties of matrix $\Wmat $. The proof can be found in the Appendix.
	
	\begin{lemma}\label{lem:1}
		Consider that all covariance matrices $\{\Cmat_1,...,\Cmat_K\}$ in the GMM are positive definite. Then, $\Wmat$ is symmetric, positive definite, and has spectral norm strictly less than 1.
	\end{lemma}

	Because the denoiser corresponds to multiplying the argument by a symmetric positive definite matrix, it is the proximity operator of a quadratic function, as stated in the following lemma.
	
	\begin{lemma}\label{lem:2}
		The function $f:\mathbb{R}^{n} \rightarrow \mathbb{R}^{n}$,	defined by $f({\bf y}) = \Wmat {\bf y}$, where $\Wmat$ is symmetric, positive definite, and has spectral norm strictly less than 1, corresponds to the proximity operator of the quadratic funtion $g({\bf y}) = \frac{1}{2}{\bf y}^T (\Wmat^{-1} - \Imat){\bf y}$.
	\end{lemma}
	
	{\noindent \bf Proof}: Using the definition of proximity operator \cite{Bauschke},
	\[
	\mbox{prox}_{g}({\bf y}) = \arg\min_{{\bf x}}  \frac{1}{2}\| {\bf x - y}\|_2^2  + \frac{1}{2}{\bf x}^T (\Wmat^{-1} - \Imat){\bf x}. 
	\]
	The function being minimized w.r.t. $\bf x$ can be written as 
	\[
	\frac{1}{2} \, {\bf x}^T \bigl(\Imat + \Wmat^{-1} - \Imat \bigr){\bf x} - {\bf x}^T{\bf y} + \frac{1}{2}\, \|{\bf y}\|_2^2,
	\]
	which is quadratic, strictly convex, with a unique minimizer $\Wmat {\bf y}$. $\blacksquare$
	
	\subsection{Convergence Proof}
	
	Lemmas~\ref{lem:1} and \ref{lem:2} allow proving convergence of the proposed algorithm, which is formalized in the following theorem.
	
	\begin{theorem}\label{th:conv}
		Consider the application of SALSA to \eqref{eq:sum1J}, which results in \eqref{eq:ineq1b} to \eqref{eq:ineq3b}. Consider also that $\bv_1, \dots, \bv_{J-1}$ are proximity operators, $\bv_J$ is obtained with the GMM-denoiser described in \ref{ssec:gmmden}, and that all the covariance matrices $\{\Cmat_1,...,\Cmat_K\}$ in the GMM are positive definite. Then, the proposed algorithm converges.
	\end{theorem}	
	
	{\noindent \bf Proof}: The proof consists in verifying the conditions of Theorem~\ref{th:admm}.
	
	The first condition for convergence is that $\Hmat$ has full column rank. The proposed algorithm is an instance of SALSA (as defined in \eqref{eq:ineq1b}--\eqref{eq:ineq3b}), which in turn is an instance of ADMM (as defined in \eqref{eq:ineq1}--\eqref{eq:ineq3}), with
	\[
	\Hmat = \begin{bmatrix} \Hmat_1 \\ \vdots \\ \Hmat_{J-1} \\ \Imat \end{bmatrix} ,
	\]
	which has full column rank, regardless of $\Hmat_1,\dots, \Hmat_{J-1}$.
	
	The second condition regards functions $g_1$ to $g_{J}$. In fact, $g_1$ to $g_{J-1}$ are closed, proper, and convex, by virtue of $\bv_1$ to $\bv_{J-1}$ being proximity operators \cite{Bauschke}. Furthermore, as shown by Lemma~\ref{lem:2}, the denoiser also corresponds to the proximity operator of a quadratic (thus also closed and proper) function $g_J$, which is strongly convex. Finally, the function being minimized is the sum of $J$ convex functions, one of which is strongly convex, thus a minimizer is guaranteed to exist. \hfill $\blacksquare$
	
\begin{figure}
			\begin{minipage}[b]{.45\linewidth}
				\centering
				\subfloat[]{\includegraphics[width=4cm]{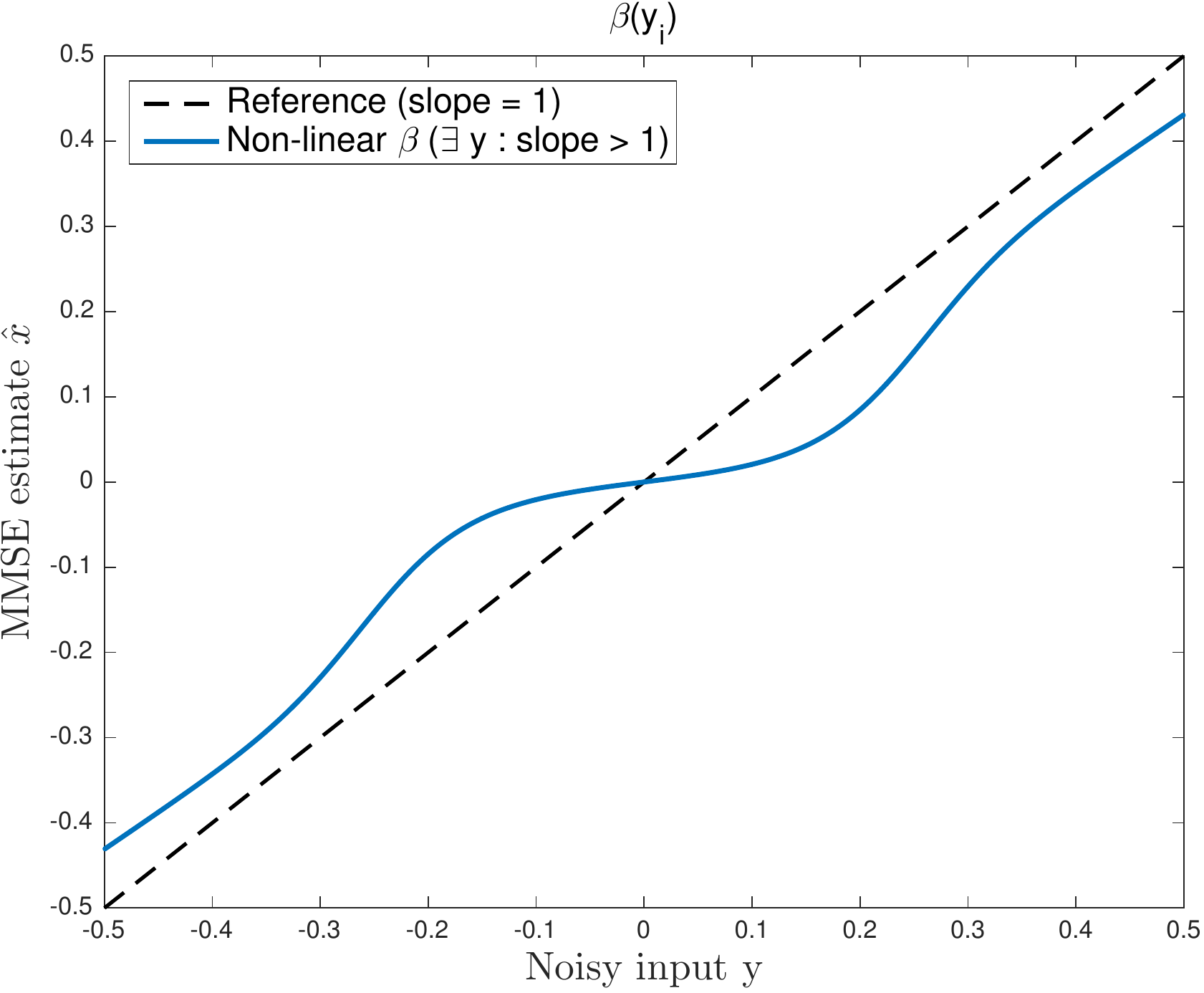}\label{fig:nonlinear}}
			\end{minipage}
			\hfill
			\begin{minipage}[b]{0.45\linewidth}
				\centering
				\subfloat[]{\includegraphics[width=4cm]{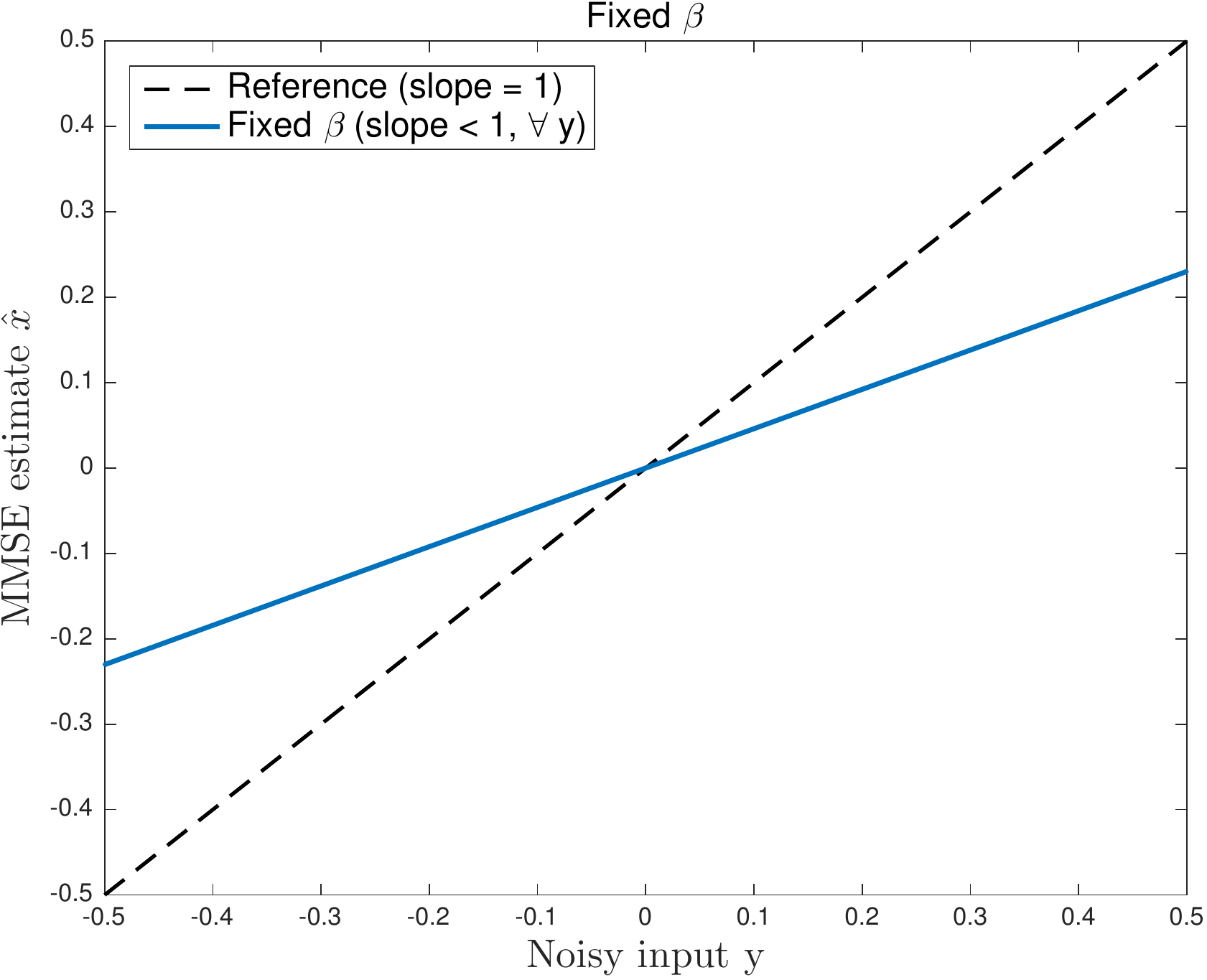}\label{fig:fixed}}
			\end{minipage}
			\caption{Denoiser expansiveness example: (a) Non-linear weights, $\beta(\by_i)$; (b) Fixed weights, $\beta$.}
			\label{fig:exp}
		\end{figure}

	\vspace{0.35cm}
	We conclude by observing that this proof of convergence depends critically on the fact that the denoiser uses fixed weights in $\bbeta$, rather than being a pure MMSE denoiser, as described in Subsection~\ref{ssec:gmmden}. In fact, the simple univariate example in Fig.~\ref{fig:nonlinear} shows that an MMSE denoiser may not even be non-expansive\footnote{Recall that a function $f:\mathbb{R}^n \rightarrow \mathbb{R}^n$ is \textit{non-expansive} if, for any $\bx,\by\in \mathbb{R}^n$, 
		\begin{equation}
		\Vert f(\bx)-f(\by) \Vert^2 \leq \Vert \bx - \by \Vert^2.
		\end{equation}
		An important connection between non-expansiveness and proximity operators in given in the following theorem:
		\vspace{-0.2cm}
		\begin{theorem}[Moreau \cite{moreau1965}]  A function is a proximal mapping if and only if it is non-expansive and the sub-gradient of some convex function.
		\end{theorem}}. This figure plots the MMSE estimate $\hat{x}$ in \eqref{eq:MMSE}, as a function of the noisy observed  $y$, under a GMM prior with two zero-mean components. Clearly, the presence of regions of the function with derivative larger than 1 shows that it is not non-expansive.

		\section{Application: Hyperspectral Sharpening}\label{sec:hs}
		
		In this section, we tackle a data fusion problem known as HS sharpening, under the PnP framework with a GMM-denoiser and a scene-adapted prior, as described above. Using compact matrix notation (as in \cite{simoes}), the HS sharpening inverse problem can be modeled as
		\begin{equation}
		\Ymat_h  =  \Zmat \Bmat \Mmat + \Nmat_h, \label{eq:yh}  \hspace{1cm} \Ymat_m = \Rmat \Zmat + \Nmat_m, 
		\end{equation}
		where $\Zmat \in \mathbb{R}^{L_h \times n_m}$ is the target image to be estimated, $\Ymat_h \in \mathbb{R}^{L_h \times n_h}$ is the observed HS data,  $\Bmat \in \mathbb{R}^{n_m \times n_m}$ is a spatial convolution operator, $\Mmat \in \mathbb{R}^{n_m \times n_h}$ is a sub-sampling operator, $\Ymat_m \in \mathbb{R}^{L_m \times n_m}$ is the observed MS data, $\Rmat \in \mathbb{R}^{L_m \times L_h}$ models the spectral responses of the MS sensor, and $\Nmat_h$ and $\Nmat_m$ are Gaussian noises of known variances. Moreover, we assume  the sensor has a fixed point spread function, and matrices  $\Bmat$ and $\Rmat$ are known; see \cite{simoes} for a blind approach. 
		
		HS data typically has hundreds of bands but the spectral vectors, i.e. the columns of $\Zmat$, live in a low dimensional subspace which may be learned from the observed HS data \cite{hysime}. In particular, for independent and identically distributed (\textit{iid}) noise, the range of $\Zmat$ may be learned from the eigenvectors of the correlation matrix $\Ymat_h\Ymat_h^T/n_h$. Then, instead of estimating $\Zmat$ directly, we estimate latent images $\Xmat \in \mathbb{R}^{L_s \times n_m}$, and recover $\Zmat = \Emat \Xmat$, where the columns of $\Emat \in \mathbb{R}^{L_h \times L_s}$ are the eigenvectors corresponding to the $L_s$ largest eigenvalues of $\Ymat_h\Ymat_h^T/n_h$. The observation model becomes
		\begin{equation} \Ymat_h  = \Emat \Xmat \Bmat \Mmat + \Nmat_h, \label{eq:yh2} \hspace{1cm} 
		\Ymat_m  =  \Rmat \Emat \Xmat + \Nmat_m. 
		\end{equation}
		
		A classical approach to this inverse problem is to compute the \textit{maximum a posteriori} (MAP) estimate, i.e., 
		\begin{align}
		\widehat{\Xmat}  \in  \underset{\Xmat}{\argmin}   \frac{\Vert \Emat \Xmat \Bmat \Mmat \! -\! \Ymat_h \Vert_F^2}{2} +  \frac{ \lambda \Vert \Rmat \Emat \Xmat \! -\! \Ymat_m \Vert_F^2}{2} + \tau \phi(\Xmat), \label{eq:optimization}
		\end{align}
		where $\phi$ is the negative log-prior (or regularizer), while $\lambda$ and $\tau$ control the relative weight of each term. We follow Sim\~oes \textit{et al.}  \cite{simoes} and use SALSA to tackle \eqref{eq:optimization}, where once again we assume the noise is Gaussian iid; more general forms of noise correlation can be included \cite{qwei}. Notice that, while in \eqref{eq:optimization} the optimization variable is a matrix $\Xmat$, in \eqref{eq:sum1J} the optimization variable is a vector $\bx$. These is merely a  difference in notation, as we can represent matrix $\Xmat$ by its vectorized version $\bx = \mbox{vec}(\Xmat)$, which corresponds to stacking the columns of $\Xmat $ into a vector. Using well-known equalities relating vectorization and Kronecker products, we can  write
		\begin{align*}
		\|  \Emat \Xmat \Bmat \Mmat  - \Ymat_h \|_F^2 & =  \left\|\bigl( \Mmat^T \otimes \Emat\bigr) (\Bmat^T \otimes \Imat) \bx - \mbox{vec} (\Ymat_h) \right\|_2^2\\
		\Vert \Rmat \Emat \Xmat  -  \Ymat_m \Vert_F^2 & =   \left\|   \bigl(\Imat \otimes (\Rmat\Emat) \bigr)  \bx - \mbox{vec} (\Ymat_m) \right\|_2^2,
		\end{align*} 
		and map \eqref{eq:optimization} into \eqref{eq:sum1J} by letting $J=3$, and
		\begin{align}
		g_1(\bv_1) &= \left\|\bigl( \Mmat^T \otimes \Emat\bigr) \bv_1 - \mbox{vec} (\Ymat_h) \right\|_2^2   \label{eq:g1} \\
		g_2(\bv_2) &=  \lambda  \left\|   \bigl(\Imat \otimes (\Rmat\Emat) \bigr) \bv_2 - \mbox{vec} (\Ymat_m) \right\|_2^2 \label{eq:g2} \\ 
		g_3(\bv_3) &=  2 \,\tau \, \phi(\Vmat_3),
		\end{align}
		where $\bv_i = \mbox{vec}(\Vmat_i)$, for $i=1,2,3$,  while 
		\begin{equation}
		\Hmat_1 = (\Bmat^T \otimes \Imat), \hspace{0.5cm}\Hmat_2 = \Imat,  \hspace{0.5cm} \Hmat_3 = \Imat.  \label{Hmatrices}
		\end{equation}
		
		Although we can arbitrarily switch between matrix and vector representations, in what follows it is more convenient to use matrix notation. Each iteration of the resulting ADMM  has the form
		\begin{align}
		& \Xmat^{k+1}  =   \arg\min_{\Xmat} \,   \Vert \Xmat \Bmat -  \Vmat^k_1  -  \Dmat^k_1 \Vert_F^2 + \Vert \Xmat  -  \Vmat_2^{k}  -  \Dmat^{k}_2 \Vert_F^2  \nonumber \\
		&  \hspace{2.2cm } + \Vert \Xmat -  \Vmat_3^{k} \! - \! \Dmat^{k}_3 \Vert_F^2,  \nonumber  \\
		& \Vmat_1^{k+1} = \arg \min_{\Vmat_1} \, \,  \Vert \Emat \Vmat_1 \Mmat - \Ymat_h \Vert_F^2 + \rho  \Vert \Xmat^{k+1}\Bmat - \Vmat_1 - \Dmat_1^{k} \Vert_F^2, \nonumber \\
		& \Vmat_2^{k+1}  =\arg \min_{\Vmat_2} \, \, \lambda\, \Vert \Rmat \Emat \Vmat_2 - \Ymat_m \Vert_F^2 + \rho  \Vert \Xmat^{k+1} - \Vmat_2 - \Dmat_2^{k} \Vert_F^2,  \nonumber \\
		& \Vmat_3^{k+1}  =\arg \min_{\Vmat_3} \, \, \phi(\Vmat_3) + \frac{\rho}{2\, \tau} \Vert \Xmat^{k+1} - \Vmat_3 - \Dmat_3^{k} \Vert_F^2, \label{eq:mpo}\\
		&\Dmat_1^{k+1} = \Dmat_1^k  + \Vmat_1^{k+1} - \Xmat^{k+1}\Bmat,\nonumber\\
		&\Dmat_2^{k+1} = \Dmat_2^k  + \Vmat_2^{k+1} - \Xmat^{k+1},\nonumber\\
		&\Dmat_3^{k+1} = \Dmat_3^k  + \Vmat_3^{k+1} - \Xmat^{k+1},\nonumber
		\end{align}
		where $\rho$ is the so-called \textit{penalty parameter} of ADMM, and $\Dmat_i$ are the scaled Lagrange dual variables \cite{boyd}. The first three problems are quadratic, thus with closed-form solutions involving matrix inversions (for details, see \cite{simoes}):
		\begin{align}
		\Xmat^{k+1} & =   
		  \left[ \left(\Vmat_1^{k} \! + \!\Dmat_1^{k}\right)\Bmat^T + \Vmat_2^{k} +  \Dmat_2^{k} + \Vmat_3^{k} +  \Dmat_3^{k} \right]\left[ \Bmat \Bmat^T  \!+ \! 2\, \Imat \right]^{-1}, \label{Xupdate} \\
		\Vmat_1^{k+1} & = \left[\Emat^T\Emat + \rho\Imat\right]^{-1}\left[\Emat^T\Ymat_h + \rho\left(\Xmat^{k+1}\Bmat - \Dmat_1^{k}\right)\right]\odot\Mmat \notag \\ & \quad + \left(\Xmat^{k+1}\Bmat - \Dmat_1^{k}\right)\odot \left(1-\Mmat\right), \label{V1update} \\
		\Vmat_2^{k+1} & = \left[\lambda\Emat^T\Rmat^T\Rmat\Emat + \rho\Imat\right]^{-1}\left[\lambda\Emat^T\Rmat^T\Ymat_m + \rho\left(\Xmat^{k+1} - \Dmat_2^{k}\right)\right], \label{V2update}
		\end{align}
		where $\odot$ denotes entry-wise (Hadamard) product.  The matrix inversion in \eqref{Xupdate} can be implemented with cost $O(n_m \log n_m)$ using the FFT, under the assumption of periodic boundary conditions. The matrix inversions in \eqref{V1update} and \eqref{V2update} involve matrices of size $L_s \times L_s$; since $L_s$ is typically $10\sim30$, the cost of this inversion in marginal. Moreover, if $\rho$ is fixed, the two inverses can be pre-computed \cite{simoes}.
		
		Once again, we tackle \eqref{eq:mpo} with the GMM-denoiser described previously, departing from the standard use of ADMM with a convex regularizer, such as the TV \cite{rudin} used in \cite{simoes}.
		
		The resulting HS sharpening formulation satisfies Theorem~\ref{th:conv}. In fact, functions $g_1$ and $g_2$ (see \eqref{eq:g1} and \eqref{eq:g2}) are quadratic, thus closed, proper, and convex. Moreover, using Lemma~\ref{lem:2}, the denoiser corresponds to the proximity operator of a quadratic function $g_3$, which is strongly convex, closed and proper. Finally, since the cost function is the sum of three convex quadratic functions, one of which is strongly convex (defined implicitly by the denoiser), a global minimum is guaranteed to exist. We stress that, although the global objective in \eqref{eq:optimization} is quadratic when $\phi$ is a GMM-based denoiser with fixed weights, and thus fixed $\Wmat$, we still need to resort to the ADMM to solve the problem because $\Wmat$ is too large to be handled directly.

		\section{Results}
		\label{sec:results}
		
		\begin{table*}[htb] 
			\begin{center}
				\caption{HS and MS fusion on (cropped) ROSIS Pavia University and Moffett Field datasets. \label{tab:sharp2}}
				\vspace{3pt}
				\resizebox{0.9\textwidth}{!}{
					\begin{tabular}{c|c||c|c|c||c|c|c||c|c|c||c|c|c}
						\multicolumn{2}{c||}{}	& \multicolumn{3}{c||}{Exp. 1 (PAN)} & \multicolumn{3}{c||}{Exp. 2 (PAN)}  & \multicolumn{3}{c||}{Exp. 3 (R,G,B,N-IR)} & \multicolumn{3}{c}{Exp. 4 (R,G,B,N-IR)} \\
						\hline 
						\hline
						Dataset & Metric & ERGAS & SAM & SRE & ERGAS & SAM & SRE & ERGAS & SAM & SRE & ERGAS & SAM & SRE\\ \hline \hline
						\multirow{3}{*}{Rosis} & Dictionary \cite{qwei} & 1.99 & 3.28 & 22.64 & 2.05 & 3.16 & 22.32 & \textbf{0.47} & \textbf{0.85} & \textbf{34.60} & 0.85 & 1.47 & 29.66 \\ \cline{2-14}
						& GMM \cite{Teodoro2017} & 1.75 & 2.89 & 23.67 & 1.92 & 2.92 & 22.85 & 0.48 & 0.87 & 34.32 & 0.91 & 1.65 & 29.05 \\ \cline{2-14}
						& \textbf{proposed GMM} & \textbf{1.65} & \textbf{2.75} & \textbf{24.17} & \textbf{1.81} & \textbf{2.76} & \textbf{23.31} & 0.49 & 0.87 & 34.59 & \textbf{0.80} & \textbf{1.42} & \textbf{30.14} \\ \hline
						\multirow{3}{*}{Moffett} & Dictionary \cite{qwei} & 2.67 & 4.18 & 20.28 & 2.74 & 4.20 & 20.05  & 1.85 & 2.72 & 23.58 & 2.12 & 3.21 & 22.25 \\ \cline{2-14}
						& GMM \cite{Teodoro2017} & 2.66 & 4.24 & 20.26 & 2.78 & 4.27 & 19.87 & 1.81 & 2.68 & 23.81 & 1.98 & 2.93 & 22.91 \\ \cline{2-14}
						& \textbf{proposed GMM} & \textbf{2.54} & \textbf{4.06} & \textbf{20.66} & \textbf{2.65} & \textbf{4.10} & \textbf{20.28} & \textbf{1.73} & \textbf{2.58} & \textbf{24.18} & \textbf{1.97} & \textbf{2.90} & \textbf{22.94} \\ \hline
						\hline
					\end{tabular}
				}
			\end{center}
		\end{table*}
		
		We compare the results using the GMM-based denoiser from \cite{Teodoro2015}, the GMM denoiser herein proposed, and the dictionary-based method from \cite{qwei} (which, to the best of our knowledge, is the state-of-the-art). We use three different metrics: ERGAS ({\em erreur relative globale adimensionnelle de synth\`ese}), SAM ({\em spectral angle mapper}), and SRE ({\em signal to reconstruction-error}) \cite{review}, in 4 different settings. In the first experiment, we consider sharpening the HS images using the PAN image, at 50dB SNR, on both the HS and PAN images. The second experiment refers to PAN-sharpening as well, with 35dB SNR on the first 43 HS bands and 30dB on the 50 remaining HS bands and the PAN image  \cite{qwei}. Experiments 3 and 4 use the same SNR as experiments 1 and 2, respectively, but the HS bands are sharpened based on 4 MS bands: R, G, B, and near-infrared. In all the experiments, a GMM with 20 components is learned from the PAN image, where the PAN and MS images are generated from the original HS bands, using IKONOS and LANDSAT spectral responses.
		
\begin{figure}[hbt]\centering
			\begin{minipage}{.25\linewidth}
				\centering
				\subfloat[]{\includegraphics[width=\textwidth]{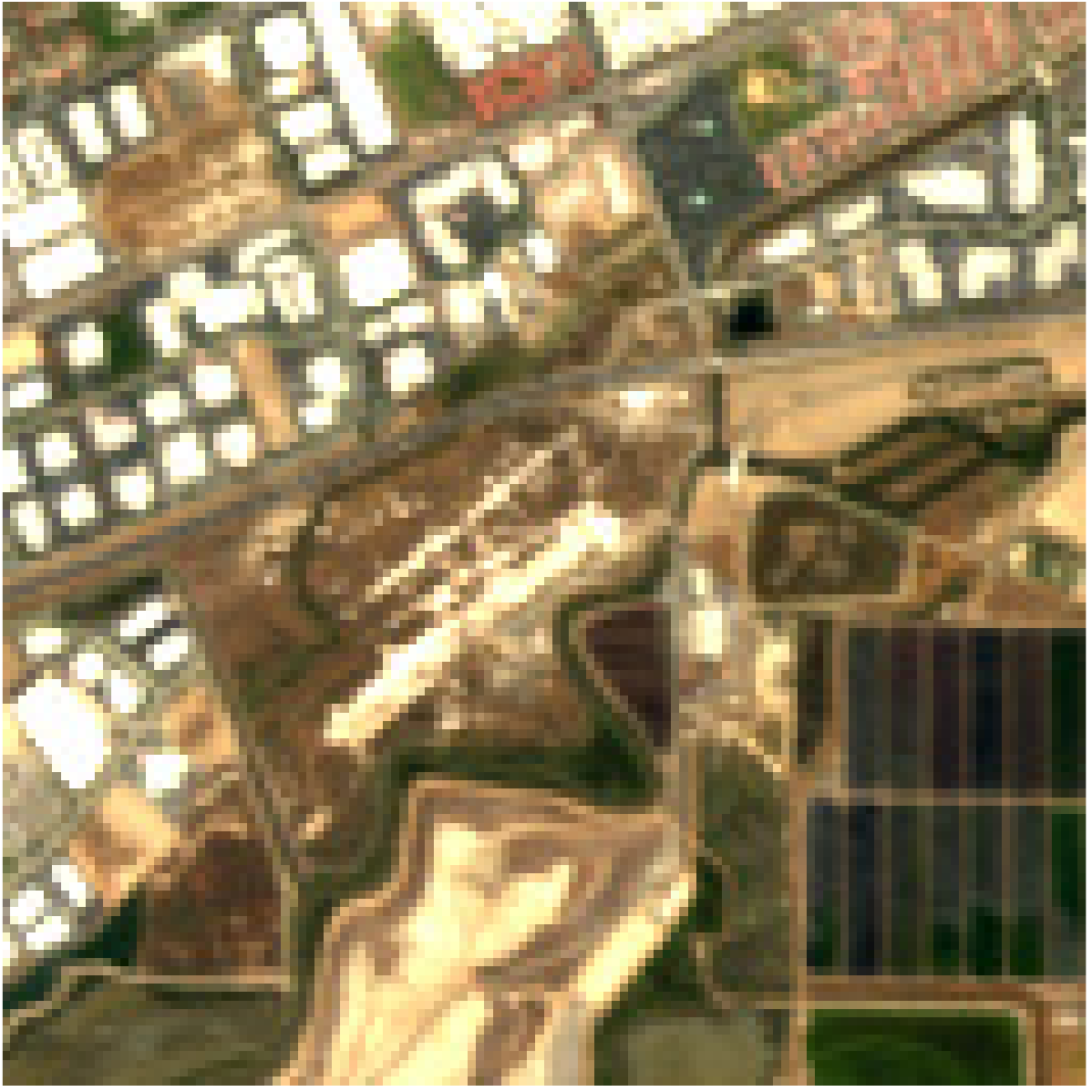}\label{fig:orig}}
			\end{minipage}%
			\begin{minipage}{.25\linewidth}
				\centering
				\subfloat[]{\includegraphics[width=\textwidth]{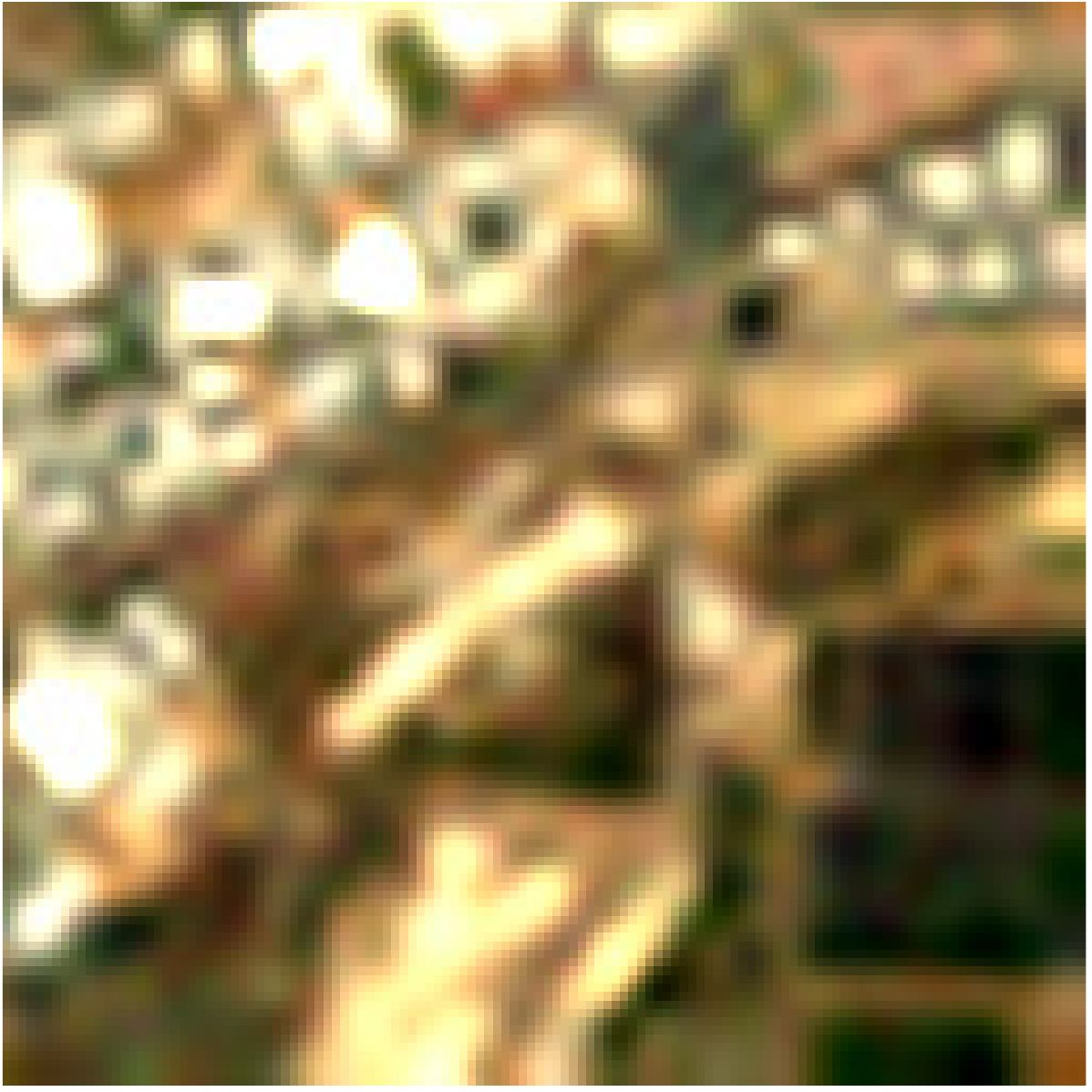}}
			\end{minipage}%
			\begin{minipage}{.25\linewidth}
				\centering
				\subfloat[]{\includegraphics[width=\textwidth]{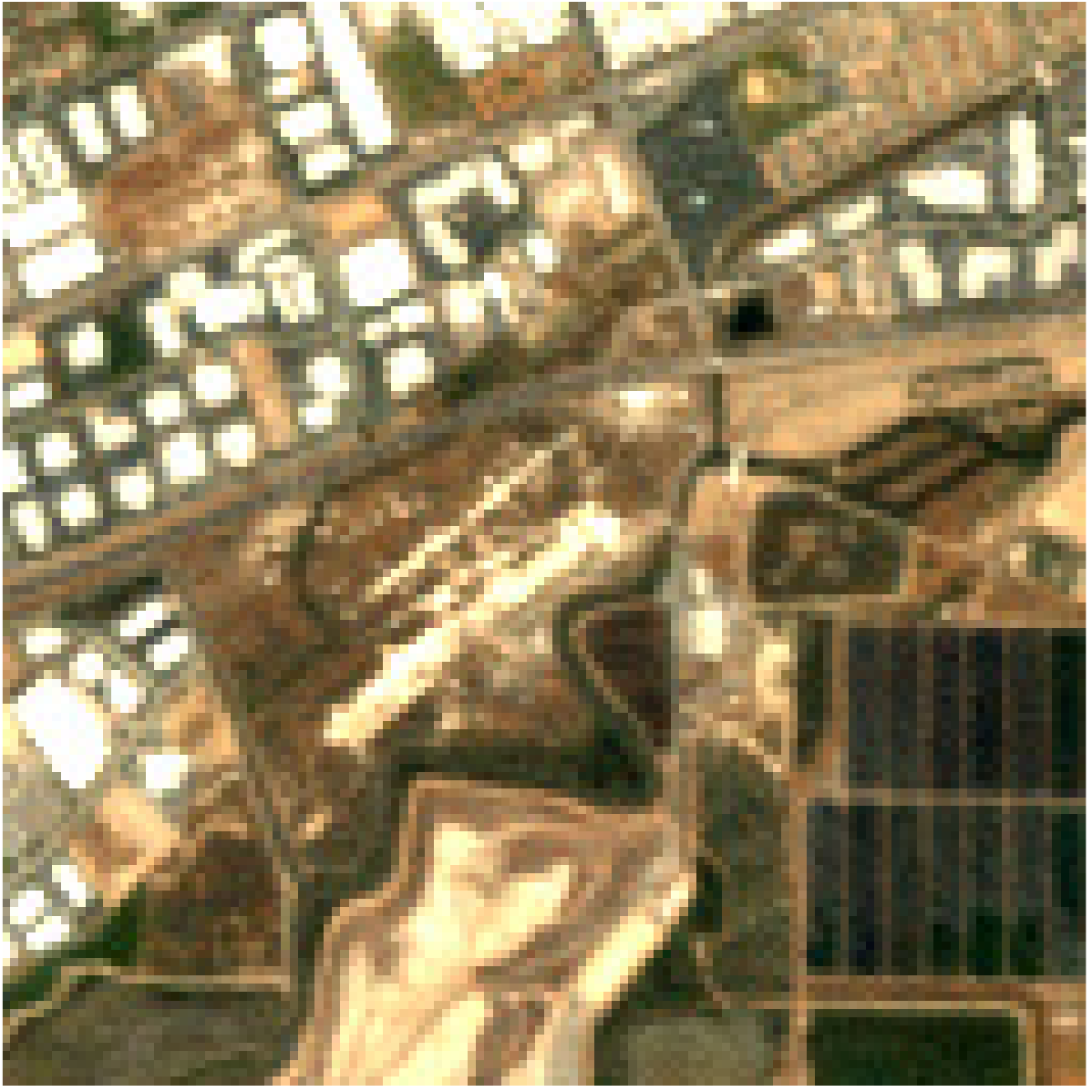}}
			\end{minipage}%
			\begin{minipage}{.25\linewidth}
				\centering
				\subfloat[]{\includegraphics[width=\textwidth]{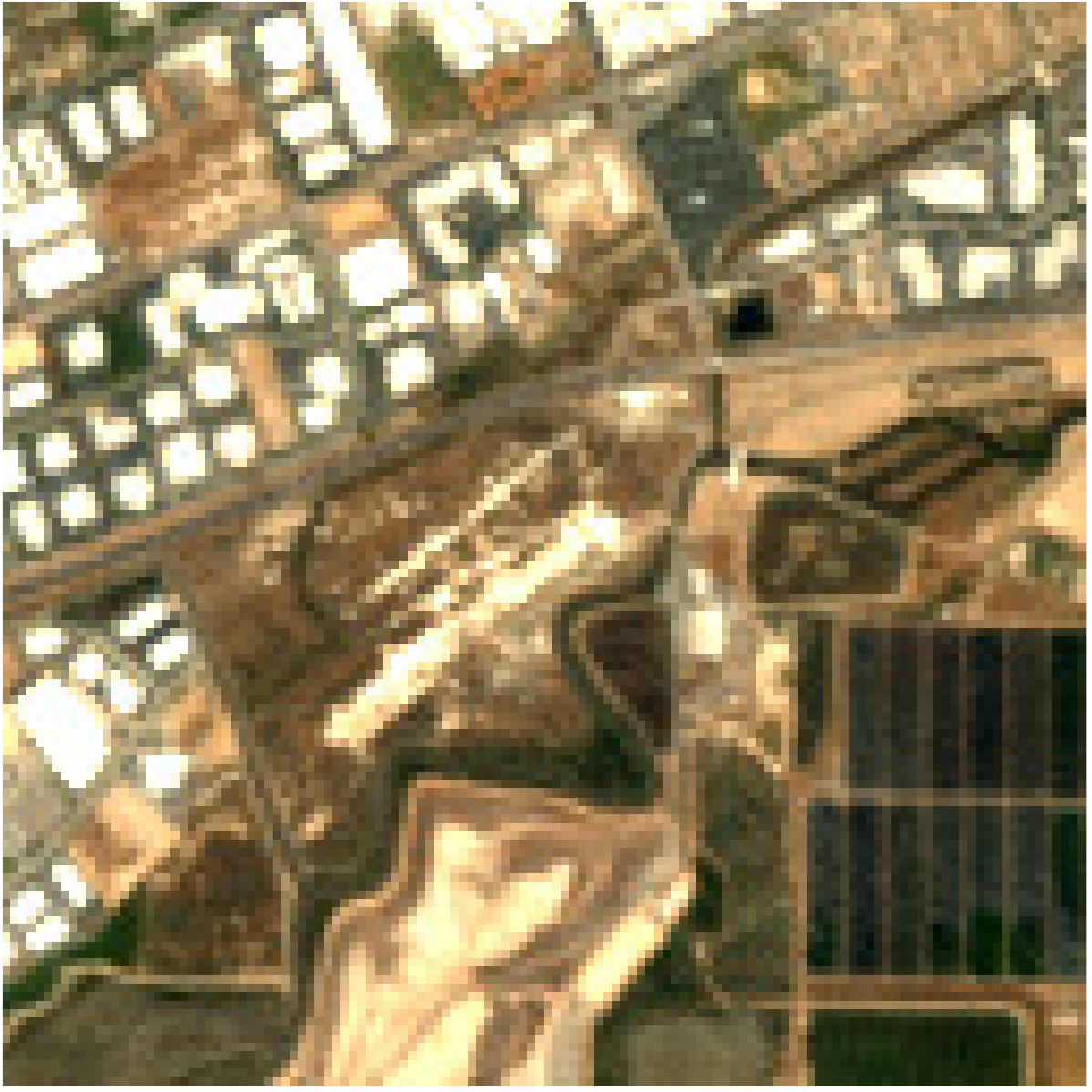}\label{fig:rec}}
			\end{minipage}
			\begin{center}
				\begin{minipage}{.33\linewidth}
					\centering
					\subfloat[]{\includegraphics[width=0.97\textwidth]{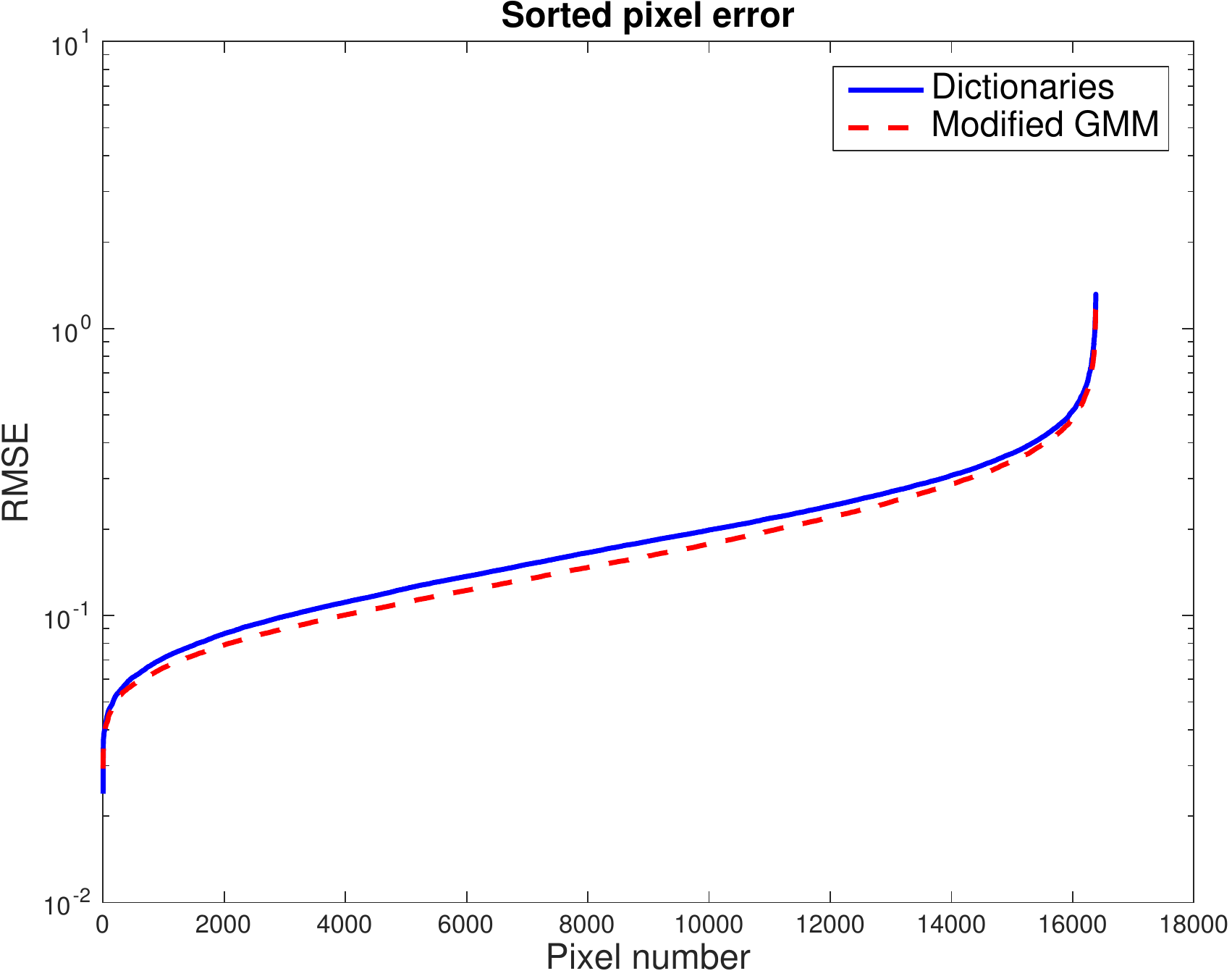}\label{fig:plotnorms2}}
				\end{minipage}%
				\begin{minipage}{.33\linewidth}
					\centering
					\subfloat[]{\includegraphics[width=0.99\textwidth]{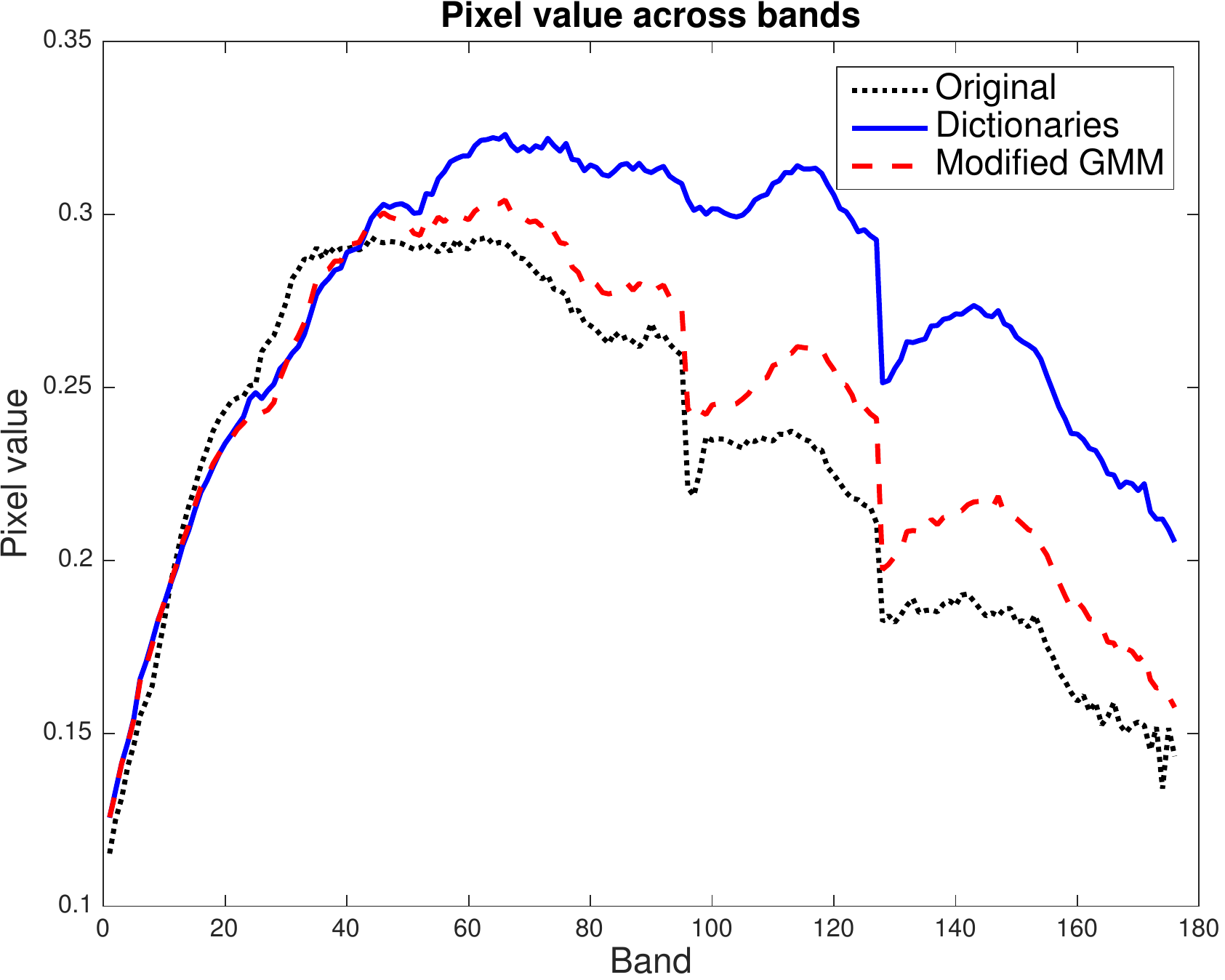}\label{fig:err1}}
				\end{minipage}%
				\begin{minipage}{.33\linewidth}
					\centering
					\subfloat[]{\includegraphics[width=0.99\textwidth]{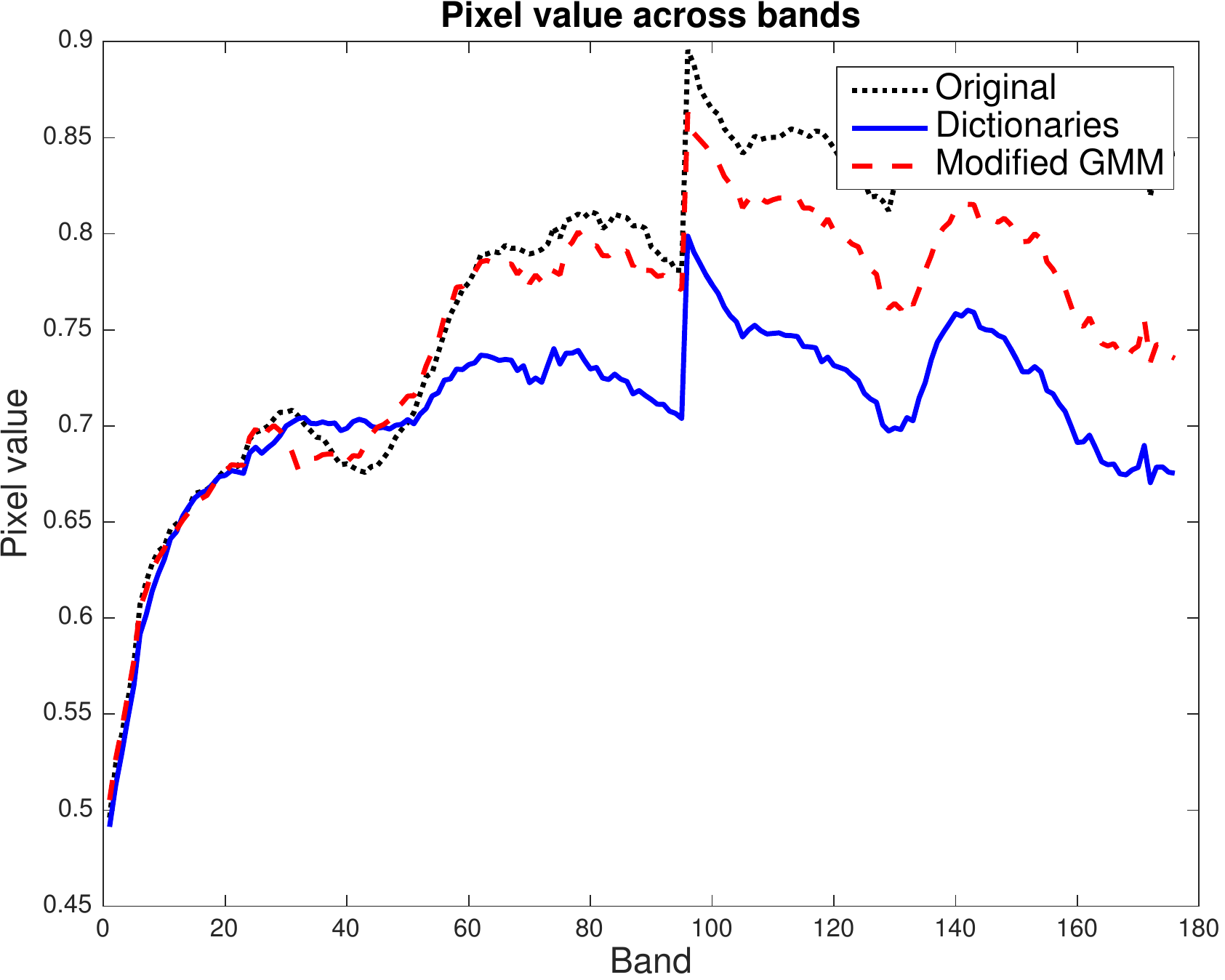}\label{fig:err2}}
				\end{minipage}
			\end{center}
			\vspace{-0.4cm}
			\caption{(a) Original HS bands in false color (20, 11, 4); (b) low-resolution; (c) dictionary-based \cite{qwei}; (d) GMM-based (proposed); (e) sorted pixel errors; (f)--(g) restored pixel value across bands.}
			\label{fig:sharp2}
		\end{figure}

		Table~\ref{tab:sharp2} shows the results on a cropped area of the \textit{ROSIS Pavia University} and \textit{Moffett Field} datasets. The three methods have comparable performances, with the proposed one being slightly better. Furthermore, the results supports the hypothesis that the target image has the same spatial structure as the PAN image used to train the GMM, otherwise keeping the posterior mixture weights would not yield good results. Figures~\ref{fig:orig}--\ref{fig:rec} show the results of experiment 4 on the Moffett Field data, in terms of visual quality, yet the differences are not noticeable. Another way to visualize the accuracy is to plot the (sorted) errors for each pixel and to notice that the proposed method has a larger number below a given threshold, \textit{i.e.}, in Figure~\ref{fig:plotnorms2} the dashed curve is below the solid one; furthermore, in Figures~\ref{fig:err1} and \ref{fig:err2}, the dashed curve is closer (smaller error) to the dotted one (the true pixel value across all bands) than the solid line.

		\section{Conclusion}
		\label{sec:conclusion}

		This paper proposed a PnP algorithm, with a GMM-based denoiser, guaranteed to converge to a global minimum of the underlying cost function. The denoiser is scene-adapted and it is the proximity operator of a closed, proper, and convex function; consequently, the standard convergence guarantees of ADMM (in Theorem \ref{th:admm}) apply, if the remaining conditions are satisfied. We illustrate the application of the algorithm on a data fusion problem known as HS sharpening.
		
		Experimental results show that the proposed method outperforms another state-of-the-art algorithm based on sparse representations on learned dictionaries \cite{qwei}, for most of the test settings. The proposed scene-adaptation also improves the results over the GMM-based denoiser in \cite{Teodoro2017}, giving supporting evidence to the hypothesis that the PAN image constitutes a useful training example, as the HS bands (and latent images) share a similar spatial structure.
		
		\section*{\smaller \smaller Appendix: Proof of Lemma~\ref{lem:1}}
		
		
		Each $\bF_i$ is symmetric, as it is a convex combination of symmetric matrices, which is clear from \eqref{eq:linMMSE} and because covariances are symmetric. Thus, $\Wmat$ is also symmetric (see \eqref{eq:defW}).
		Consider the eigendecomposition of each ${\bf C}_j = \bU_j^T \bSigma_j \bU_j$,  where $\bSigma_j = \mbox{diag}(\varsigma_1^j,...,\varsigma_{n_p}^j)$ contains its eigenvalues,  in non-increasing order. Then,
		\begin{equation}
		{\bf C}_j \Bigl(  {\bf C}_j + \sigma^2 \; \Imat \Bigr)^{-1} =  \bU_j^T \, \bSigma_j \left( \bSigma_j + \sigma^2 \; \Imat \right)^{-1} \bU_j,
		\end{equation}
		where $\bSigma_j \left( \bSigma_j + \sigma^2\; \Imat \right)^{-1}$ is a diagonal matrix, and thus the eigenvalues  of ${\bf C}_j (  {\bf C}_j + \sigma^2\; \Imat )^{-1}$ are in 
		\[
		\bigl[ \varsigma_{n_p}^j / (\varsigma_{n_p}^j + \sigma^2),\;  \varsigma_1^j/ (\varsigma_1^j + \sigma^2) \bigr] \in (0,1),  
		\]
		since $\varsigma_{n_p}^j > 0$ ($\Cmat_j$ is positive definite) and $\sigma^2 > 0$. 
		
		From \eqref{eq:linMMSE}, each $\bF_i$ is a convex combination of matrices, each of which with eigenvalues in $(0,1)$. Weyl's inequality \cite{Bhatia} implies that the eigenvalues of a convex combination of symmetric matrices is bounded below (above) by the same convex combination of the smallest (largest) eigenvalues of those matrices. The eigenvalues of $\bF_i$ are thus all in $(0,1)$, \textit{i.e.}, $\bF_i$ is positive definite and $\|\bF_i\|_2 < 1$.
		
		Finally, following \cite{Sulam}, we partition the set of patches into a collection of subsets of non-overlapping patches: $\{\Omega_j \subset \{1,..,,N\}, \; j=1,...,n_p \}$ (the number of subsets of non-overlapping patches is equal to the patch size, due to the assumption of unit stride and periodic boundaries). Using this partition, $\Wmat$ can be written as
		\begin{align}
		\Wmat =  \frac{1}{n_p} \sum_{j=1}^{n_p} \underbrace{\sum_{k \in \Omega_j} \bP_k^T \bF_k \bP_k}_{\Amat_j} = \frac{1}{n_p} \sum_{j=1}^{n_p} \Amat_j. \label{eq:noover}
		\end{align}
		Since the patches in $\Omega_j$ are disjoint, there is a permutation of the image pixels that allows writing $\Amat_j$ as a block-diagonal matrix, where the blocks are the $\bF_k$ matrices, with $k \in \Omega_j$. Because the set of eigenvalues of a block-diagonal matrix is the union of the sets of eigenvalues of its blocks, the eigenvalues of each $\Amat_j$ are bounded similarly as those of the $\bF_k$, thus in $(0,1)$. Finally, again using Weyl's inequality, the eigenvalues of $\Wmat$ are bounded above (below) by the average of the largest (smallest) eigenvalues of each $\Amat_j$, thus also in $(0,1)$. \hfill $\blacksquare$

		\bibliographystyle{IEEEbib}
		\bibliography{refs}

\end{document}